\documentclass[sigconf]{acmart}
\AtBeginDocument{%
  \providecommand\BibTeX{{%
    \normalfont B\kern-0.5em{\scshape i\kern-0.25em b}\kern-0.8em\TeX}}}

\copyrightyear{2022} 
\acmYear{2022} 
\setcopyright{acmcopyright}\acmConference[NLBSE'22]{The 1st Intl. Workshop on Natural Language-based Software Engineering}{May 21, 2022}{Pittsburgh, PA, USA}
\acmBooktitle{The 1st Intl. Workshop on Natural Language-based Software Engineering (NLBSE'22), May 21, 2022, Pittsburgh, PA, USA}
\acmPrice{15.00}
\acmDOI{10.1145/3528588.3528653}
\acmISBN{978-1-4503-9343-0/22/05}

\usepackage{url}
\usepackage{tabularx}
\usepackage{todonotes}
\usepackage{cancel}

\begin{document}

\title{Can NMT Understand Me? Towards Perturbation-based Evaluation of NMT Models for Code Generation}

\author{Pietro Liguori}
%\email{webmaster@marysville-ohio.com}
\affiliation{%
  \institution{University of Naples Federico II}
  %\streetaddress{P.O. Box 1212}
  \city{Naples}
  \country{Italy}
}
\email{pietro.liguori@unina.it}
\orcid{0000-0001-5579-1696}

\author{Cristina Improta}
\affiliation{%
  \institution{University of Naples Federico II}
  %\streetaddress{P.O. Box 1212}
  \city{Naples}
  \country{Italy}
}
\email{crist.improta@studenti.unina.it}

\author{Simona De Vivo}
\affiliation{%
  \institution{University of Naples Federico II}
  %\streetaddress{P.O. Box 1212}
  \city{Naples}
  \country{Italy}
}
\email{simona.devivo@unina.it}

\author{Roberto Natella}
\affiliation{%
  \institution{University of Naples Federico II}
  %\streetaddress{P.O. Box 1212}
  \city{Naples}
  \country{Italy}
}
\email{roberto.natella@unina.it}

\author{Bojan Cukic}
\affiliation{%
  \institution{University of North Carolina at Charlotte}
  %\streetaddress{P.O. Box 1212}
  \city{Charlotte}
  \state{North Carolina}
  \country{USA}
}
\email{bcukic@uncc.edu}

\author{Domenico Cotroneo}
\affiliation{%
  \institution{University of Naples Federico II}
  %\streetaddress{P.O. Box 1212}
  \city{Naples}
  \country{Italy}
}
\email{cotroneo@unina.it}

\renewcommand{\shortauthors}{Liguori et al.}

\begin{abstract}
Neural Machine Translation (NMT) has reached a level of maturity to be recognized as the premier method for the translation between different languages and aroused interest in different research areas, including software engineering. A key step to validate the robustness of the NMT models consists in evaluating the performance of the models on adversarial inputs, i.e., inputs obtained from the original ones by adding small amounts of perturbation. 
However, when dealing with the specific task of the code generation (i.e., the generation of code starting from a description in natural language), it has not yet been defined an approach to validate the robustness of the NMT models.
In this work, we address the problem by identifying a set of perturbations and metrics tailored for the robustness assessment of such models.
We present a preliminary experimental evaluation, showing what type of perturbations affect the model the most and deriving useful insights for future directions.
\end{abstract}

\begin{CCSXML}
<ccs2012>
   <concept>
       <concept_id>10010147.10010178.10010179.10010180</concept_id>
       <concept_desc>Computing methodologies~Machine translation</concept_desc>
       <concept_significance>300</concept_significance>
       </concept>
 </ccs2012>
\end{CCSXML}

\ccsdesc[300]{Computing methodologies~Machine translation}
%\keywords{datasets, neural networks, gaze detection, text tagging}
\keywords{neural machine translation, robustness testing, code generation, adversarial inputs}
%%  \includegraphics[width=\textwidth]{sampleteaser}
%%  \caption{Seattle Mariners at Spring Training, 2010.}
%%  \Description{Enjoying the baseball game from the third-base
%%  seats. Ichiro Suzuki preparing to bat.}
%% \end{teaserfigure}

\maketitle

\section{Introduction}
\label{sec:introduction}
%Topic: code generation or attack generation
%Additional information?
%https://inst.eecs.berkeley.edu/~cs161/fa08/papers/stack_smashing.pdf
As in many areas of artificial intelligence, deep neural networks have become the dominant paradigm for machine translation, bringing impressive improvements in the quality of the translation, and continuously moving forward the state-of-the-art performance~\cite{koehn2020neural}. 
%but also new challenges.

Unlike traditional phrase-based translation, which consists of many small sub-components tuned separately, Neural Machine Translation (NMT) attempts to build and train a single, large neural network that reads a sentence and outputs a correct translation~\cite{bahdanau2014neural}. 
NMT has reached a level of maturity to be recognized as the premier method for the translation between different languages~\cite{wu2016google} and aroused interest in different research areas, including software engineering. 
In particular, the \textit{code generation} task, also lately referred to as \textit{semantic parsing}~\cite{yin2019reranking,xu2020incorporating}, is an emerging and important application of NMT. It consists in the automatic translation of an \textit{intent} in natural language (NL), such as the English language, into a \textit{code snippet} written in a specific programming language. 
Indeed, NMT has been extensively used for generating programs (e.g.,  Python~\cite{yin2017syntactic} and Java~\cite{ling2016latent}), or to perform other programming tasks, such as code completion \cite{drosos2020wrex,shi2020tf}, the generation of UNIX commands~\cite{lin2017program,lin2018nl2bash}, etc.
%The \textit{code generation} task, also lately referred to as \textit{semantic parsing}~\cite{yin2019reranking,xu2020incorporating}, has been previously addressed for several programming languages, e.g. Python~\cite{yin2017syntactic} and Java~\cite{ling2016latent} code snippets. 
%NMT has also been used to perform other programming task, such as code completion \cite{drosos2020wrex,shi2020tf}, the generation of UNIX commands~\cite{lin2017program,lin2018nl2bash} or commit messages~\cite{jiang2017automatically,liu2018neural,jung2021commitbert}, etc.
Recently, NMT techniques have been also adopted to automatically generate code for software exploits starting from the description in natural language~\cite{liguori2021shellcode_ia32,liguori2021evil,liguori2022generate}. 

%To feed the NMT models for the generation of code, researchers build corpora containing a huge amount of translation from the intents into the code snippets. 
A common situation in any translation task from NL to programming language is the gap between the natural language used in the corpora and the natural language actually used by programmers. 
As a matter of fact, the corpora used for NMT models are often too ``literal'' and cumbersome to be realistically used by programmers. 
For example, in the \textit{Shellcode\_IA32} dataset~\cite{liguori2021shellcode_ia32,liguori2022generate} used for the generation of assembly code from natural language, the \textit{intent}, i.e., the natural language description, \textit{``Push the contents of eax onto the stack''} takes longer than writing the assembly instruction ``\texttt{push eax}''. The \textit{Django} dataset~\cite{oda2015learning}, which is widely used for evaluating neural machine translation task from English to Python \cite{yin2018mining,hayati2018retrieval,yin2018tranx}, contains numerous Python code snippets that are relatively short (e.g., ``\texttt{chunk\_buffer = BytesIO(chunk)}'') described with with English statements that are definitely longer than the snippets (\textit{``evaluate the function BytesIO with argument chunk, substitute it for chunk\_buffer.''}). Again, in the \textit{CoNaLa} dataset~\cite{yin2018mining}, we can find shortcode snippets (e.g., ``\texttt{GRAVITY = 9.8}'') described with longer English intents (\textit{``assign float 9.8 to variable GRAVITY''}). 

%Since different users express the English intents in their own way, a user may state the input omitting information implicitly contained in the intent, or may use synonyms not actually used in the training data. 

Since different users express the English intents in their own way, NMT models need to be robust against gaps between the actual intents and the ones in the corpora. 
A key approach typically used in machine learning research is to perform \emph{robustness testing} of models, i.e., to evaluate the performance of the models when dealing with unexpected inputs, and to identify cases of misclassification. In particular, robustness testing has been adopted to identify security issues in machine learning models, by crafting \textit{adversarial inputs}~\cite{zhang2021crafting}, i.e., inputs obtained from the original ones by adding small amounts of perturbation, which a malicious attacker may generate to mislead the model. 
% -> computer vision - esempio di sicurezza
% -> cinese - esempio di correttezza funzionale
%
These kinds of attacks on the inputs were first investigated for computer vision systems. 
Recent studies also addressed this problem in the context of language translation (e.g., from English to Chinese) by injecting noise in the input at different linguistic levels~\cite{belinkov2018synthetic,li2018textbugger,huang2021robustness}.

Given the gap discussed above, NMT models may not be robust to intents that are valid descriptions of the code, but that follow different styles or have different levels of detail compared to the training corpus. If NMT models are unable to handle this variability, they would be too inflexible and hamper the productivity of the programmers, hence limiting their usability in practice. 
Therefore, to evaluate the robustness of the NMT models, we aim to introduce non-arbitrary perturbations, e.g., variations from well-intentioned users. 
This is still an open research problem: while images can be easily perturbed without losing their original meaning and semantics, perturbing natural language can be much more challenging.

In light of these considerations, our work provides three key contributions:
\begin{itemize}
    \item We propose a set of perturbations to evaluate the robustness of NMT models for the code generation task. The set includes both perturbations already used in previous studies and identified as suitable for the code generation task, and novel ad-hoc perturbations for the code generation task;
    \item We identify a set of metrics tailored for the robustness evaluation of NMT models under different levels of perturbations. Indeed, a significant aspect to take into account is that, while a perturbed intent may produce an output different from the original one, it may still preserve the semantic and syntactic correctness according to the target programming language;
    \item We present a preliminary experimental analysis to evaluate the robustness of an NMT model when dealing with perturbations in the intents. We show what perturbations affect the model the most and derive useful insights for future research.
\end{itemize}

In the following, Section~\ref{sec:related} discusses related work; Section~\ref{sec:perturbation} proposes a set of perturbations to evaluate the robustness of NMT models; Section~\ref{sec:metrics} describes the metrics for the evaluation of the model robustness; Section~\ref{sec:evaluation} presents the preliminary evaluation;  Section~\ref{sec:conclusion} concludes the paper. 

\section{Related Work}
\label{sec:related}
State-of-the-art provides several recent works on adversarial natural language processing (NLP) covering different research topics such as sentiment analysis, toxic content detection, machine comprehension, and numerous similar contexts.
%There is much recent work on adversarial NLP, especially focused on machine translation tasks, sentiment analysis, toxic content detection, machine comprehension, and numerous similar contexts.

%While DNNs models have proven to be capable to achieve great results in the aforementioned domains and many others, such as image recognition and classification, recently they have been found to be vulnerable against adversarial examples, as in carefully crafted inputs obtained from the original ones by adding small amounts of perturbation to mislead the classification.
%These sort of attacks were first investigated for computer vision systems; however, an important role in this adversarial setting is played by the task and its specific inputs: while image space is continuous, and as such can be easily slightly modified without it being perceivable, attacking text strings, “intents”, without altering the original meaning and semantics can be much more challenging.

%Previous studies explored and analyzed noise generation at different linguistic levels.
%Recent studies addressed the problem in the context of language translation (e.g., from English to Chinese) by injecting noise in the input to evaluate the robustness of the NMT models.
Previous works explored and analyzed noise generation at different linguistic levels, i.e., character, word, and sentence-level.
At character-level, text can be perturbed by inserting, deleting, randomizing, or swapping characters to study the effects on natural language processing (NLP) tasks \cite{heigold2018robust,belinkov2018synthetic,li2018textbugger}; furthermore, homographic attacks can be employed to mislead models in question answering \cite{wu2020evaluating}, and QWERTY character swapping can be used to reproduce keyboard typos \cite{belinkov2018synthetic}.
At the word level, words in a sentence can be substituted with different random words, similar words in the word embedding space, or meaning-preserving words \cite{li2018textbugger,michel2019evaluation,huang2021robustness}.
Regarding sentence-level manipulation, paraphrasing, back translation, and reordering are some of the approaches used to produce a syntactically and semantically similar phrase to fool the models~\cite{huang2021robustness}.

%\vspace{1pt}
%\noindent
%\textbf{Machine translation.} 
Heigold \textit{et al.}~\cite{heigold2018robust} studied the effects of word scrambling and random noise insertion in NLP tasks such as morphological tagging and machine translation, both regarding English and German languages. The perturbation strategies used include character flips and swaps of neighboring characters to imitate typos.
Belinkov \textit{et al.} \cite{belinkov2018synthetic} analyzed how natural noise, i.e., the natural occurring of errors from available corpora, and synthetic noise, i.e., character swaps aimed to reproduce misspellings and keyboard typos, affect character-based NMT models, focusing on machine translation from natural languages such as French, German and Czech to English. The authors used a black-box adversarial training setting and found that these architectures have a tendency to break when presented with noisy datasets.%but seeing them during training can make them more robust. 

%\vspace{1pt}
%\noindent
%\textbf{Machine comprehension.}
In the context of the machine comprehension and question answering, Wu \textit{et al.}~\cite{wu2020evaluating} investigated what type of text perturbation leads to the most high-confidence misclassifications and which embeddings are more susceptible to adversarial attacks. They used homographic attacks, synonyms substitutions, and sentence paraphrasing to investigate models' performances in a perturbed context paragraph.
%\vspace{1pt}
%\noindent
%\textbf{Semantic parsing.} 
Huang \textit{et al.} \cite{huang2021robustness} conducted the first empirical study to evaluate the effect of adversarial examples on SOTA neural semantic parsers by perturbing existing benchmark corpora with four different word-level operations and two sentence-level operations and applying meaning-preserving constraints.%The authors' ultimate goal is to provide a benchmark for evaluating both standard accuracy and robustness metrics. 

%\vspace{1pt}
%\noindent
%\textbf{Sentiment analysis.} 
Recent works introduced tools and frameworks for the generation of adversarial inputs.
TextBugger~\cite{li2018textbugger} is a framework to efficiently generate utility-preserving adversarial texts under both white-box and black-box settings to evaluate the robustness of various popular, real-world online text classification systems. In the white-box scenario, the attacker is aware of the model’s architecture and parameters, so they first find important words by computing the Jacobian matrix of the classifier, then choose an optimal perturbation from the generated five kinds of perturbations. In the black-box scenario, the attacker does not have information on the model’s internals, so they first find the significant sentences and then use a scoring function to find the main words to manipulate. Specifically, their targets are sentiment analysis and toxic contents detection models.  %Ultimately, they demonstrate that TextBugger is able to exploit vulnerabilities of sentiment analysis and toxic contents detection models and fool the classifiers with a high success rate.
Gao \textit{et al.} \cite{gao2018blackbox} presented DeepWordBug, an algorithm to effectively generate small text perturbations in a black-box setting. The authors use novel scoring and ranking techniques to identify the most important words that, if perturbed, lead the model to a misclassification. Concerning these perturbations, they apply character-level transformations such as swap, substitution, deletion, and insertion. 
Cheng \textit{et al.}~\cite{cheng2020seq2sick} proposed Seq2Sick, an optimization-based framework to generate adversarial examples for sequence-to-sequence neural network models. The authors implemented novel loss functions to conduct a non-overlapping attack and targeted keyword attack, to handle the almost infinite output space.

Our work can be considered complementary to the previous ones. Indeed, although the robustness evaluation of the deep learning models has been widely addressed by the previous research, to the best of our knowledge, the use of adversarial attacks has not been applied to validate the usability of the NMT models in the code generation task.
%While research is abundant regarding robustness to noise in machine translation tasks from one natural language to another, to our knowledge, adversarial attack and training techniques have not been applied in the code generation domain.
%In this context, a significant aspect to take into account is that, while a perturbed intent may produce an output different from the original one, it may still be a syntactically correct statement in the target programming language.

\section{Perturbations in Code Generation}
\label{sec:perturbation}
To measure the robustness of the NMT models in the code generation task, we are interested to analyze the models with respect to their inputs (i.e., intents in natural language). Indeed, the description of a natural language code snippet by different authors may be characterized by different writing styles and capabilities. A sentence may be rephrased through multiple synonyms, it may order words in different ways, it may lack some significant detail, or be too specific. 

Therefore, although character-level perturbations may be meaningful to study the sensitivity of NMT models to human errors (e.g., typos), in this work we focus on perturbing words in a sentence but still preserving the original meaning of the intents. In particular, we focus on two types of perturbations: the \textit{unseen synonyms}, and the \textit{missing information}. The former can be used to evaluate the performance of the translation task when the intents significantly diverge from the terms used in the corpus (e.g., word synonyms). 
The latter, instead, is suitable to assess the models' performance when programmers may omit information that would be redundant, such as information implicitly contained in the sentence, or information already stated in previous intents. Both these aspects are important for the usability of NMT models.

%\vspace{2pt}
%\noindent
%\textbf{Word Synonyms.} 
\subsection{Unseen Synonyms}
A robust model should be resistant to noise caused by Unseen Synonyms and should produce the same output when presented with two semantically similar intents. Therefore, it is interesting to our cause to substitute words within an intent either with a synonym from a lexical database (e.g., WordNet~\cite{miller2015wordnet}) or with their neighbor in the \textit{word embedding} space (i.e., a numerical representation of the words)~\cite{mikolov2013linguistic} and examine the model’s response. 

However, blindly replacing words with their synonym may lead to the loss of the sentence’s original meaning since terms with small word embedding distance may belong to the same context but not be semantically similar (e.g., the words ``\textit{father}'' and ``\textit{mother}''). Moreover, code generation is a highly specific domain, thus some words have a precise meaning and cannot simply be replaced with another. 
%As simple example, consider the intent ``\textit{call the function}''. A valid perturbation on the input can reasonably lead to the sentence ``\textit{execute the function}'', but not to ``\textit{phone the function}'' since the verb phone is clearly out of the programming context. 
As a simple example, consider the intent ``\textit{clear the contents of the register}''. A valid perturbation on the input can reasonably lead to the sentence ``\textit{empty the contents of the register}'', but not to ``\textit{purify the contents of the register}'' since the verb ``\textit{purify}'' is clearly out of the programming context.

To overcome these issues, a solution could be limiting the space of the possible words by creating a dictionary of words used to describe programming code (e.g., by using books and tutorials as reference). However, building a vocabulary from scratch containing only words used in the programming language context may be too time-consuming or, even worse, unfeasible. 
A more practical approach consists in applying \textit{constraints} on the transformation method. An example of constraints for synonyms is to ensure that the words can be replaced only with one of its top k-nearest neighbors in the source embedding space before computing a similarity score to filter out dissimilar terms~\cite{li2018textbugger,huang2021robustness}.

The use of the constraints for the choice of synonyms also allows limiting situations in which the new word produces a different meaning from the original intent. Referring to the previous example ``\textit{clear the contents of the register}'', a synonym without constraints for the verb ``\textit{clear}'' is the verb ``\textit{shift}''~\cite{oxford}, which is definitely used in the programming code context, but with a completely different purpose.
Taking this into account, we identified three different types of constraints useful to perform word substitution in the intents: 
\begin{itemize}
    \item \textit{\textbf{Word Embedding Distance}}: It measures the value of the cosine similarity between word embeddings. %A word embedding can be described as a real-valued vector that encodes the meaning of a word such that words that are closer in the vector space are expected to have similar meaning \cite{mikolov2013linguistic}. 
    The constraint-based on the word embedding distance performs the substitution of words only if the value of the cosine similarity between the replaced word and its synonym is higher than a specified value;
    \item \textit{\textbf{BERT-score}}: It measures token similarity between two texts using \textit{contextual embedding}~\cite{zhang2020bertscore}. Contextual embeddings, such as BERT, can generate different vector representations for the same word in different sentences depending on the surrounding words, which form the context of the target word~\cite{devlin2019bert}. By using the constraint on the BERT-score, the substitution of the words is performed only if the score between the replaced word and its synonym is higher than a specified value;
    \item \textit{\textbf{Part-of-Speech (POS) tag}}: It is the process of marking up a word in a text as corresponding to a particular part of speech. The constraint using the POS tag allows the substitutions only if the replaced word and its synonym have the same POS tag (e.g., a verb should be replaced only with a verb, a noun with a noun, etc.).
\end{itemize}

%Taking this into account, we only apply a transformation if the chosen WordNet synonym and the original word have a BERT-score similarity \cite{zhang2020bertscore} greater than 0.85 and the same Part-of-Speech tag.

\begin{comment}
\tablename{}~\ref{tab:unseen} shows how the use of unseen synonyms work on the original intent ...
\begin{table}[ht]
\centering
\caption{Examples of unseen synonyms in the intents.}
\label{tab:unseen}
\small
\begin{tabular}
{ >{\centering\arraybackslash}m{2cm} |   >{\centering\arraybackslash}m{5cm}}
\toprule
\textbf{\textit{Original Intent}}                    & define the decoder function and store the encoded shellcode pointer in the esi register\\\midrule
\textbf{\textit{Unseen Synonyms w/o constraints}}    & \textit{fix} the decoder function and store the encoded shellcode pointer in the esi register\\\midrule
\textbf{\textit{Unseen Synonyms with constraints}}   & \textit{determine} the decoder function and store the encoded shellcode pointer in the esi register\\ \bottomrule
\end{tabular}
\end{table}
\end{comment}

%\vspace{2pt}
%\noindent
%\textbf{Word Removal.} 
\subsection{Missing Information}
In the context of code generation, the removal of information becomes of particular interest since the intents of the corpora are usually concise and detailed, thus they may completely lose their original meaning even if only a single word is omitted or removed. Nevertheless, this represents a common situation because users can inadvertently neglect some details, or avoid specifying information implicitly contained in the intent or included in the previous ones.

The action of removing information from the intents can be performed randomly~\cite{huang2021robustness} or following particular criteria. In our case, it is interesting to analyze how the model’s behavior and text comprehension varies when important information is missing.  This kind of perturbation is yet to be explored in the code generation task. For this reason, we first define what \textit{important information} means in our context before removing one or more significant words from each intent.
When commenting on a code snippet, there are two fundamental aspects to be considered: i) \textit{what action the user aims to take}, and ii) \textit{what is the target of the action}. For example, the simple intent ``\textit{call the myfunc function}'' contains the action, i.e., the verb \textit{call}, and the target, i.e., the \textit{myfunc function}. The target of the action can be further divided in the value of the target (i.e., the name of the function), and the word specifying the type (i.e., the word ``function'').
%Moreover, the aforementioned target can be further distinguished in two sets of different terms: words specific to the programming language, such as keywords, and words specific to the underlying architecture, for instance, registers’ names.
Based on these assumptions, we identify three main categories of significant words in the intents: 
\begin{itemize}
    \item \textit{\textbf{Action-related words}}: Words containing the information related to the actions of the intent, which are usually specified by the verbs (e.g., jump, add, call, declare, etc.);  
    \item \textit{\textbf{Language-related words}}: Words related to the target programming language (e.g., the words ``class'', ``function'', ``variable'', ``register'', ``label'', etc.);
    \item \textit{\textbf{Value-related words}}: They include the name or the values of the variables, the names of functions, classes and, where available (e.g., assembly language), the value of the memory addresses, and of name of registers or labels. 
\end{itemize}

\tablename{}~\ref{tab:missing} shows the different types of word removal perturbations on the English intent ``\textit{Store the shellcode pointer in the ESI register.}'', which is commonly used to decode shellcodes in assembly language for the IA-32 architecture~\cite{liguori2021evil}.
The table shows examples in which the intent still preserves its meaning even without specifying the omitted words. The verb store and the keyword register are implicit (the pointer of the shellcode can be only moved to \texttt{ESI}, which is, in fact, a register), while the name of the register can be derived from the context of the program (the \texttt{ESI} register is commonly used to store the shellcode). 
However, this is not always the case. For example, a list can be created or deleted, therefore, not specifying the verb can imply an opposite action. A user can create non-primitive data structures, but the type of the structure (e.g., list, dictionary, etc.) has to be specified to perform a correct prediction. Finally, values and names have a broader range of meaning and usage, hence it might be more difficult for a model to learn and predict their behavior.

\begin{table}[t]
\centering
\caption{Examples of omitted information on the same intent. \textcolor{red}{\textbf{\cancel{Slashed}}} text refers to the omitted words.}
\label{tab:missing}
%\small
\begin{tabular}
{ >{\centering\arraybackslash}m{3cm} |   >{\centering\arraybackslash}m{4cm}}
\toprule
\textbf{Perturbation}           &   \textbf{Intent}\\ \midrule
\textit{None (Original Intent)} &   Store the shellcode pointer in the ESI register.\\ \midrule
\textit{Action-related words}   &   \textcolor{red}{\textbf{\cancel{Store}}} the shellcode pointer in the ESI register.\\ \midrule
\textit{Language-related words} &   Store the shellcode pointer in the ESI \textcolor{red}{\textbf{\cancel{register}}}.\\  \midrule
\textit{Value-related words}    &   Store the shellcode pointer in the \textcolor{red}{\textbf{\cancel{ESI}}} register.\\ \bottomrule
\end{tabular}
\end{table}

%Consequently, our focus is on understanding how much missing information a model can tolerate without too much performance degradation as well as understanding which category’s removal affects performances the most and why.
%A preliminary observation can be done simply by analyzing the nature of programming languages and their natural language description. Some target words usually require the same verb: functions can be defined, labels can be defined, lists can be created, and so on. Therefore, even in their absence, the output can still be predicted. 
%On the contrary, value-related information has a broader range of meaning and usage, hence it might be more difficult for a model to learn and predict their behavior. 

\section{Evaluation Metrics}
\label{sec:metrics}
%The evaluation of the robustness of NMT models for code generation is not trivial in that we need to take into account different aspects.
%When the input is subject to perturbations, we need to understand if the output predicted by the model is \textit{correct}, i.e., it is equivalent to the reference of the test set (i.e., the ground truth).
%In the context of programming code generation, the correctness can be syntactic and semantic. The former gives insights into whether the predicted code is correct from the point of view of the language syntax (e.g., it would be compilable), the latter indicates whether the predicted output is the exact translation of the intent into the target programming language.
When the input is perturbed, we need to assess if the output predicted by the model is \textit{correct}, i.e., it is equivalent to the reference of the test set (i.e., the \textit{ground-truth}).
However, the robustness evaluation of the NMT models is not trivial in that we need to take into account different aspects.

The \textit{ambiguity of the natural language} implies that the same sentence can have different meanings and, therefore, it can be translated into different and non-equivalent programming code snippets. This problem is further exacerbated by the introduction of perturbations on the intents (e.g., word synonyms, omitted words, etc.). 
A significant takeaway is that, although the model's prediction can be incorrect with respect to the reference, it can result in the right translation of the perturbed intent.

As well as in natural language we can express the same intents with different sentences (e.g., through the use of synonyms, sentence paraphrases, etc.), the \textit{equivalence of code snippets} allows programmers to write different but equivalent programming code. This means that, even if the output predicted by the model differs from the ground truth, it can still be considered correct. 

In the light of the above considerations, the choice of the right metrics is a key step to assess the robustness of the NMT models in the code generation task.
In the remainder of this section, we describe a set of metrics suitable for this specific research problem.

\subsection{Automatic Metrics}
Automatic metrics are a valuable means to assess the quality of the code generation task since they are reproducible, easy to be tuned, and time-saving.
%Despite they do not credit correct code prediction that fails to match the reference of the test set, these metrics are a valuable means to assess the quality of the code generation task since they are reproducible, easy to be tuned, and time-saving.

Among the most commonly used metrics in machine translation, we definitely find the \textbf{\textit{Bilingual Evaluation Understudy}} (BLEU) score and the \textbf{\textit{Exact Match Accuracy}} (EM)~\cite{ling2016latent,yin2017syntactic,tran2019does,gemmell2020relevance,yin2018tranx,yin2019reranking,al2022survey}. 
BLEU score~\cite{papineni2002bleu} is based on the concept of \textit{n-gram}, i.e., the adjacent sequence of $n$ \textit{items} (e.g., syllables, letters, words, etc.) from a given example of text or speech. This metric measures the degree of n-gram overlapping between the strings of words produced by the model and the references at the corpus level. BLEU measures translation quality by the accuracy of translating n-grams to n-grams, for n-gram of size $1$ to $4$~\cite{han2021translation}.
The Exact Match Accuracy, instead, measures the fraction of the exact match between the output predicted by the model and the reference in the test set.
%These two metrics are not ideal: accuracy only measures exact match and cannot thus give credit to semantically correct code that is different from the reference, while it is not clear whether BLEU provides an appropriate proxy for measuring semantics in the code generation task~\cite{yin2017syntactic}.

%BLEU is an algorithm that automatically measures the quality of machine translation by evaluating the correspondence between machine output and human output. In addition, BLEU can calculate the similarity between the translation of the original sentence and the translation of a variation of the same sentence~\cite{wei2020retrieve,sun2020automatic}. The Exact Match Accuracy, instead, represents the fraction of the exact match between the output predicted by the model and the reference in the test set.
%These metrics well fits in the context of the code generation since the former evaluates the similarity between the translation of the perturbed sentence and that of the original sentence, the latter evaluates the correct output only if it perfectly matches the prediction. 

Further metrics useful in the context of the robustness evaluation are based on sub-string analysis~\cite{sun2020automatic}. For example, the \textbf{\textit{LCS-based metric}} measures the normalized similarity by calculating the longest common sub-sequence between the translation to the output of the original input and the translation to the output of the mutated input, respectively. The \textbf{\textit{Ed-based metric}} measures the edit-distance between two strings, where edit-distance is a way of quantifying dissimilarity between two strings (i.e., the minimum number of operations required to make two strings equal).

\subsection{Manual Metrics}
Although automatic metrics can evaluate the differences between the output predicted by the model and the reference of the test set, the automatic evaluation can not truly reflect the \textit{correctness} of the predicted code when it differs from the reference of the test set~\cite{stent2005evaluating}. 
Therefore, to properly assess the robustness of the models, we need to evaluate the quality of the code snippets by using manual metrics, i.e., metrics that are computed through human inspection. 
%These metrics are time-consuming and prone to errors but allow us to assess the deeper linguistic features of the language~\cite{han2021translation}.
%However, to properly evaluate the model's prediction, it is important to define the concept of correctness. 
In the context of the code generation task, in order to estimate the correctness of the output, we need to look into the code with respect to i) \textit{how the code is written}, i.e., the \textit{code syntax} and ii) \textit{what the code actually does}, i.e., the \textit{code semantic}. 

Therefore, a key step to evaluate the correctness of the model's output is to estimate both the \textbf{\textit{Syntactic Accuracy}} and \textbf{\textit{Semantic Accuracy}} (also \textbf{\textit{Execution Accuracy}})~\cite{wang2018robust,liguori2021evil}, which measure the fraction of syntactic and semantic correct predictions over all the predictions, respectively. 
While the former gives insights into whether the code is correct according to the rules of the target language, the latter indicates whether the output is the exact translation of the intent into the target programming language.
The semantic correctness implies syntax correctness, while a snippet can be syntactically correct but semantically incorrect. Of course, the syntactic incorrectness also implies the semantic one~\cite{liguori2021evil}. 

Different from the syntax, the evaluation of the code snippet semantic depends by definition on the intent considered as reference.
For example, semantic accuracy assesses if the prediction, after the perturbation, is correct according to the intent of the original test set.
%Since we aim to evaluate the robustness of the NMT models through adversarial examples, the evaluation of the semantic can be performed also with respect to the perturbed intent~\cite{huang2021robustness}.
%Huang \textit{et al.}~\cite{huang2021robustness} proposed a set of metrics tailored to estimate the robustness of the NMT models.
The \textbf{\textit{Perturbation Accuracy}}~\cite{huang2021robustness}, instead, computes the fraction of predictions considered correct with respect to the perturbed version of the intents, i.e., the output is considered correct if it is the exact translation of the perturbed input into the target programming language.
This metric is of particular interest when the perturbation introduces ambiguity or, even worse, changes the semantic meaning of the intent. In this case, indeed, the output may be considered correct according to the perturbed version of the intent but incorrect when considering the intent in the original test set (not perturbed) as the reference, and vice-versa. 

A further metric of interest in this context is the \textbf{\textit{Robust Accuracy}}~\cite{huang2021robustness}. The metric focuses the evaluation on the intents of the original test set which are properly predicted by the model without any perturbations, discarding the ones mispredicted by the model. To assess the robustness, it computes the fraction of correct predictions under perturbations over the subset of the previous correct outputs. The metric is based on the assumption that to evaluate the model's robustness, it may be meaningless to include intents leading to the model's mispredictions, regardless of the perturbations.

\section{Preliminary Evaluation}
\label{sec:evaluation}
We performed a set of preliminary experiments to assess the model’s ability to tolerate noise and still produce accurate outputs.
We targeted the Seq2Seq model since it is widely used in a variety of neural machine translation tasks. In particular, we adopted the Seq2Seq model with Bahdanau-style attention mechanism~\cite{bahdanau2014neural}.
We implemented the Seq2Seq model using {\fontfamily{qcr}\selectfont xnmt} \cite{neubig2018xnmt}. We used an Adam optimizer \cite{kingma2015adam} with $\beta_1=0.9$ and $\beta_2=0.999$, while the learning rate $\alpha$ is set to $0.001$. We set all the remaining hyper-parameters in a basic configuration: layer dimension = $512$, layers = $1$, epochs (with early stopping enforced) = $200$, beam size = $5$. 
We did not use any pre-processing or post-processing steps to help the model in the generation of the output since we are interested in quantifying the impact of the noise rather than maximizing the performance.

To feed the model, we used the assembly dataset released by Liguori \textit{et al.}~\cite{liguori2021evil} for automatically generating assembly from natural language descriptions. 
This dataset consists of assembly instructions, commented in English language, which were collected from shellcodes for \textit{IA-32} and written for the \textit{Netwide Assembler} (NASM) for Linux \cite{duntemann2000assembly}.
The dataset contains $3,715$ unique pairs of assembly code snippets/English intents: $3105$ pairs in the training set, $305$ pairs in the dev set, and $305$ pairs in the test set.

Our preliminary evaluation interested a subset of the perturbations described in \S{}~\ref{sec:perturbation}. In particular, we evaluated the robustness of the model by using three different types of perturbations: 
\begin{itemize}
    \item \textit{Unseen synonyms with constraints using the BERT-score and POS tag}: We applied a transformation only when the synonym, chosen as a neighbor in the word embedding space, and the original word have a BERT-score similarity greater than $0.85$ and the same POS tag. We empirically choose a high value for the BERT-score similarity to introduce diversity in the intent without losing the original meaning. We randomly replaced the 10\% of the selected words within a single intent, ensuring that at least one word is swapped with its synonym in each intent.
    \item \textit{Omission of the action-related words}: We removed the verbs from every intent in the test set using a POS tagger (e.g., “define”, “add”, etc.);
    \item \textit{Omission of the language-related words}: We removed the words related to the assembly programming language from each intent (e.g., ``register'', ``label'', etc.) in the test set.
\end{itemize}

We used TextAttack~\cite{morris2020textattack}, a Python framework for data augmentation in NLP, to replace words with synonyms and apply the constraints, and Flair POS-tagging model~\cite{akbik2018coling} as part-of-the speech tagger.
The TextAttack framework implements the \textit{word swap by embedding} transformation, i.e., a novel \textit{counter-fitting} method for injecting linguistic constraints into word vector space representations, which post-processes word vectors to improve their usefulness for tasks involving the semantic similarity judgements~\cite{mrksic:2016:naacl}.

We perturbed all the intents of the test set (i.e., the test set is 100\% perturbed), while we did not add any noise in the training and dev sets. All experiments were performed on a Linux OS running on a virtual machine with 8 CPU cores and 8 GB RAM. 

\subsection{Automatic Evaluation}

\begin{comment}
\begin{table}[t]
\centering
\caption{Automatic evaluation of different types of adversarial inputs. The worst performance is \textcolor{red}{\textbf{red/bold}}.}
\label{tab:automatic}
\begin{tabular}
{ >{\centering\arraybackslash}m{4cm} |   >{\centering\arraybackslash}m{1.75cm} >{\centering\arraybackslash}m{1.25cm} }
\toprule
\textbf{Test Set} & \textbf{BLEU-4 (\%)} & \textbf{EM (\%)} \\ \toprule
\textit{Original (no perturbations)}        & 17.39 & 19.67 \\ 
\textit{Unseen synonyms with const.}        & 16.03 & 18.11 \\
\textit{Action-related words}               & 13.45 & \textcolor{red}{\textbf{13.11}}\\
\textit{Language-related words}             & \textcolor{red}{\textbf{13.09}} & 16.39\\
\bottomrule
\end{tabular}
\end{table}
\end{comment}

% Tabella aggiornata con le altre metriche automatiche

%\begin{comment}
\begin{table*}[t]
\centering
\caption{Automatic evaluation of different types of adversarial inputs. The worst performance is \textcolor{red}{\textbf{red/bold}}.}
\label{tab:automatic}
%\resizebox{\columnwidth}{!}{
\begin{tabular}
{ >{\centering\arraybackslash}m{4cm} |   >{\centering\arraybackslash}m{2cm} >{\centering\arraybackslash}m{2cm} >{\centering\arraybackslash}m{2cm} >{\centering\arraybackslash}m{2cm} }
\toprule
\textbf{Test Set} & \textbf{BLEU-4 (\%)} & \textbf{EM (\%)}  & \textbf{ED (\%)}   & \textbf{LCS (\%)} \\ \toprule
\textit{Original (no perturbations)}        & 17.39 & 19.67 & 62.48 & 64.70 \\ 
\textit{Unseen synonyms with const.}        & 16.03 & 18.11 & 59.53 & 62.66 \\
\textit{Action-related words}               & 13.45 & \textcolor{red}{\textbf{13.11}} & \textcolor{red}{\textbf{53.19}} & \textcolor{red}{\textbf{56.08}} \\
\textit{Language-related words}             & \textcolor{red}{\textbf{13.09}} & 16.39 & 56.09 & 58.48 \\
\bottomrule
\end{tabular}
%}
\end{table*}
%\end{comment}

%To measure the robustness of the model, we used both automatic and manual metrics. In particular, we adopted the BLEU-4 and the exact match accuracy (ACC) as automatic metrics and evaluated the syntactic and semantic correctness of each code snippet predicted by the model. 
%\tablename{}~\ref{tab:results} shows the performance of the code generation task when different types of noise are applied on the intents. 
We first evaluated the performance of the code generation task in terms of automatic metrics both on the original and on the perturbed test set. The key idea is that, the more the performance decreases compared to the one of the original test set, the more the model is affected by the perturbation. As automatic metrics, we used the BLEU-4, the exact match accuracy (EM), the Ed-based metric (ED), and the LCS-based metric (LCS). \tablename{}~\ref{tab:automatic} shows the results. 

Among the type of perturbations, the use of unseen synonyms with constraints less affect the performance of the model. The model, indeed, showed to be robust when dealing with word synonyms, also because the high BERT-score similarity set as constraint limited the amount of diversity of the words. 
The explicit information removal from the intents, instead, negatively impacted the model's prediction. In particular, the removal of the action-related words implied the worst performance in terms of exact match accuracy, Ed-based metric, and the LCS-based metric, while the model shows the worst BLEU-4 when dealing with the removal of the language-related words.

\subsection{Manual Evaluation}

The previous metrics do not provide a complete and robust evaluation: EX only measures exact match and cannot thus give credit to semantically correct code that is different from the reference, while it is not clear whether BLEU provides an appropriate proxy for measuring semantics in the code generation task~\cite{yin2017syntactic}.
Therefore, we further studied the impact of perturbations on the code generation task by performing a manual evaluation. In particular, for each code snippet predicted by the model, all authors evaluated both the syntactic and semantic accuracy, independently. To reduce the possibility of errors in the manual analysis, multiple authors discussed cases of discrepancy, obtaining a consensus for the syntactic and semantic correctness.
 \tablename{}~\ref{tab:manual} shows the percentage of syntactically (SYN) and semantically (SEM) correct snippets over all the examples of the test set. 

The table shows that the use of perturbations does not negatively impact the model's ability to predict syntactically correct code snippets. Even better, the removal of action-related words slightly increased the performance of the syntactical accuracy of the model. Through an in-depth analysis of the model's outputs, we found that the removal of verbs resulted in the prediction of relatively simple code snippets (in terms of length) and, thus, syntactically correct, but which do not represent the exact translation of the original intent. As a matter of fact, the removal of the action-related words resulted in the most significant dropping of the performance in terms of semantic accuracy.
Similarly, the use of unseen synonyms and the removal of language-related words negatively affected the semantic accuracy of the model, but the dropping of semantic accuracy is more limited. In particular, the table shows that the semantic accuracy of the outputs achieved when the language-related words are omitted is close to the one of the original test set. 

We conducted a \textit{paired-sample T-test} to compare the syntactic and the semantic accuracy values of the code snippets predicted under perturbations with the ones of the original test set (given the same example). 
We found that the differences in the syntactic accuracy obtained under different types of perturbations are not statistically significant from the one of the original test set. Concerning the semantic accuracy, the hypothesis testing suggested that the performance achieved with the use of unseen synonyms and the removal of the action-related words are statistically significant with $p < 0.01$. The difference of the performance achieved with the removal of the language-related words, instead, did not result in any statistical evidence.
%We found that the differences between CodeBERT and the Seq2Seq are statistically significant for both metrics with $p < 0.05$.
%For the assembly programs, Seq2Seq provides a higher percentage of syntactically correct snippets, but these differences are not statistically significant. Again, CodeBERT outperforms Seq2Seq in the semantic correctness ($p < 0.05$).

\begin{table}[t]
\centering
\caption{Manual evaluation of different types of adversarial inputs. The worst performance is in \textcolor{red}{\textbf{red/bold}} ($*=$ p$<$0.01).}
\label{tab:manual}
\begin{tabular}
{ >{\centering\arraybackslash}m{4cm} |   >{\centering\arraybackslash}m{1.5cm} >{\centering\arraybackslash}m{1.5cm}  }
\toprule
\textbf{Test Set} & \textbf{SYN (\%)} & \textbf{SEM (\%)} \\ \toprule
\textit{Original (no perturbations)}    & 88.52  & 22.95\\ 
\textit{Unseen synonyms with const.}    & \textcolor{red}{\textbf{87.87}}  & 18.36*\\
\textit{Action-related words}           & 89.51	 & \textcolor{red}{\textbf{14.75}}*\\
\textit{Language-related words}         & 88.20	 & 20.98\\
\bottomrule
\end{tabular}
\end{table}

A significant takeaway from this preliminary evaluation is that in the generation of assembly code from natural language, the NMT model:
 i) can deal with the use of synonyms in the intents and, therefore, different ways of describing the code by different users; 
ii) is very robust to non-explicit information on language-related words, such as keywords; 
iii) is hugely affected by intents where actions are non explicitly stated.

\begin{comment}
\subsection{Qualitative Analysis}
\label{subsec:qualitative}

\begin{table*}[ht]
\centering
\caption{Illustrative examples of correct and incorrect output for unseen synonyms and omitted information. The prediction errors are \textcolor{red}{\textbf{red/bold}}.}
\label{tab:cases_unseen}
\begin{tabular}{
 >{\centering\arraybackslash}m{5cm}|
 >{\centering\arraybackslash}m{5cm}|
 >{\centering\arraybackslash}m{5cm}}

\toprule
\textbf{Intent Perturbation} & \textbf{NL Intent}  & \textbf{Model Output}\\ \midrule
\textit{None (Original Test Set)} & & \\
\textit{Unseen Synonyms with constraints}  & & \\
\textit{Action-related words}  & & \\ 
\textit{Language-related words}  & & \\\midrule
\textit{None (Original Test Set)} & & \\
\textit{Unseen Synonyms with constraints}  & & \\
\textit{Action-related words}  & & \\ 
\textit{Language-related words} & & \\
\bottomrule
\end{tabular}
\end{table*}
\end{comment}

\section{Conclusion and Future Work}
\label{sec:conclusion}
We addressed the problem of evaluating the robustness of the NMT models for the code generation task by proposing a set of perturbations and metrics to assess the impact of the models when dealing with different inputs.
We performed a preliminary evaluation of the Seq2Seq model in the assembly code generation from natural language description and showed how different perturbations on the inputs affect the model's performance.

As future work, we aim to extend the robustness evaluation to different DL-based architectures~\cite{sun2020treegen,feng2020codebert}. 
We are also investigating different solutions to make NMT models more robust. In particular, we foresee the use of the \textit{adversarial training} (i.e., injecting perturbed inputs into training data to increase robustness)~\cite{ebrahimi2018adversarial,cheng2019robust,ji2020adversarial} and the development of solutions that help the models to derive the missing or implicit information from the context of the program~\cite{tiedemann2017neural,wang2017exploiting,agrawal2018contextual,scherrer2019analysis}.

\begin{acks}
This work has been partially supported by the University of Naples Federico II in the frame of the Programme F.R.A., project id OSTAGE.
\end{acks}

\bibliographystyle{ACM-Reference-Format}
\balance
\bibliography{biblio}   % sample-base -> biblio

%\appendix

\end{document}